\documentclass[10pt,twocolumn]{ICCAS}
\usepackage{amsmath}
\usepackage{amssymb} 
\usepackage{graphicx}
\usepackage{subcaption}
\usepackage{caption}
\usepackage{lipsum}
\usepackage{tikz}
\usepackage{url}
\usetikzlibrary{arrows.meta,positioning,fit,calc,shapes.geometric}
\usepackage{diagbox}

\begin{document}

\title{ Terminal Constraint Model Predictive Control for Image-Based \\Visual Servoing of UAVs with Kalman Filter–Based Moment Loss Compensation}

\author{X.Wang${}^{1*}$, Y.Cao${}^{1}$, W.L.W.Leong${}^{1}$,Y.R.Tan${}^{1}$, S.Huang${}^{1}$,  S.H.R.Teo${}^{1}$ and C.Xiang${}^{1}$ }

\affils{ ${}^{1*}$College of Design and Engineering, National University of Singapore, Singapore, \\
\{e1350382, e1323388\}@u.nus.edu, \{william.leong, yanrui, tslhs, tsltshr, e1exc\}@nus.edu.sg   
  {\small${}^{*}$ Corresponding author}
 }

\abstract{
Image-Based Visual Servoing (IBVS) provides an efficient vision-guided control paradigm for unmanned aerial vehicles (UAVs) by directly regulating image-space errors. However,  conventional IBVS controllers are vulnerable to two critical issue: loss of closed-loop stability near the target due to input and state constraints, and control failure caused by intermittent loss of moment-based visual features under aggressive motion. To address these challenges, this paper proposes a terminal-constraint model predictive control (TC-MPC) framework for IBVS, integrating with a Kalman filter (KF)– based state-prediction mechanism. The TC-MPC explicitly incorporates terminal-state constraints and a terminal cost into the IBVS error dynamics, ensuring recursive feasibility, improved convergence behavior, and closed-loop stability under control and state constraints. In parallel, the Kalman filter predicts the temporal evolution of image moments during short-term visual degradation, enabling the controller to preserve control continuity when moment measurements are partially unavailable. The proposed approach is validated through real-time UAV visual servoing experiments.
}

\keywords{
    Kalman filter,model predictive control, unmanned aerial vehicle, visual servoing control.
}

\maketitle


\section{Introduction}
Drone control systems typically rely on global positioning systems (GPS) for navigation \cite{ref0}. However, GPS signals are often unreliable or unavailable in indoor, forested, and dense urban environments due to signal attenuation and obstruction. Visual servoing offers an alternative solution by utilizing visual features extracted from onboard cameras for feedback control. Unlike conventional navigational or localization approaches \cite{ref1}, \cite{ref2}, \cite{ref3}, visual servoing does not require explicit 3D reconstruction of the target pose or the drone’s full 3D pose. Instead, it directly incorporates 2D image features into the control loop, enabling the drone to adjust its position and orientation such that the observed features align with desired target locations in the image plane. This approach is computationally efficient and comparatively simple to implement, making it well-suited for navigation and positioning tasks in complex or uncertain environments.

In general, visual servoing approaches are broadly categorized into Position-Based Visual Servoing (PBVS) and Image-Based Visual Servoing (IBVS). PBVS estimates the 3D pose of the target and reconstructs the environment in Cartesian space to generate control commands. In contrast, IBVS directly utilizes image-plane information for feedback control without requiring explicit 3D reconstruction of either the target or the environment. Instead, 2D image features extracted from onboard cameras are incorporated directly into the control loop, allowing the UAV to regulate its position and orientation such that the observed image features converge to desired locations in the image plane. Owing to its relatively low computational complexity, robustness to calibration errors, and direct coupling between perception and control, IBVS is particularly suitable for navigation and positioning tasks in complex or uncertain environments. Consequently, IBVS has been widely adopted in UAV applications such as target tracking, autonomous landing, and navigation through cluttered environments \cite{ref4}, \cite{ref5}.

IBVS approaches can be categorized into three main classes: classical IBVS, spherical projection-based IBVS, and image moment-based IBVS. Classical IBVS methods directly utilize image features extracted from pixel coordinates to formulate the control law \cite{ref6,ref7,ref8}. While these approaches are conceptually simple and widely adopted, they may suffer from singularities and poor camera trajectories under certain configurations \cite{ref9}. To address these limitations, spherical projection models have been introduced to derive alternative IBVS control laws by representing visual features on a unit sphere \cite{ref10,ref11}. Although this formulation can alleviate some singularity-related issues, it often results in more complicated image kinematics and controller design. In contrast, IBVS employs image moments as visual features for visual servoing control \cite{ref12,ref13,ref14,ref170}. Image moment-based methods can simplify the associated kinematic and dynamic formulations while providing robustness to image perturbations. However, most existing works do not explicitly consider practical system constraints, such as actuator saturation and velocity limits. In \cite{ref15}, a model predictive control (MPC)-based framework was proposed to incorporate input constraints into IBVS. Nevertheless, the method was developed for classical IBVS and may therefore still encounter singularity issues. Furthermore, the proposed approach was validated only in simulation without experimental flight demonstrations.

Therefore, this work proposes an improved Image-Based Visual Servoing (IBVS) framework that integrates Model Predictive Control (MPC) with Kalman Filter for robust visual servoing control. The proposed approach first derives the dynamic characteristics of image moments and subsequently formulates an MPC-based IBVS controller that explicitly accounts for both state and control constraints. To evaluate the effectiveness of the proposed method, the controller is deployed on a quadrotor UAV for autonomous gate traversal tasks. In contrast to prior UAV gate navigation approaches based on spherical projection models \cite{ref16}, the proposed framework leverages image moment features within an MPC formulation to achieve constrained visual servoing while maintaining a comparatively simpler image representation and control structure.

\section{Image moment-based MPC}
\subsection{Moment-based visual servoing control}\label{moment-based}

Image moments are scalar values calculated from an image's pixel intensities, providing useful information about the image’s shape and distribution.

For a grayscale image $I(x,y)$, the $(p,q)$-th order \textit{moment} is defined as
\begin{equation}
M_{pq} = \sum_x \sum_y x^p y^q I(x,y),
\end{equation}
where $p$ and $q$ denote the orders of the moment, $x$ and $y$ are the pixel coordinates, and $I(x,y)$ represents the pixel intensity at location $(x,y)$.

Central moments are computed with respect to the image centroid $(\bar{x}, \bar{y})$ and are defined as
\begin{equation}
\mu_{pq} = \sum_x \sum_y (x - \bar{x})^p (y - \bar{y})^q I(x,y),
\end{equation}
where the centroid coordinates are given by
\begin{equation}
\bar{x} = \frac{M_{10}}{M_{00}}, \quad
\bar{y} = \frac{M_{01}}{M_{00}}.
\end{equation}
Therefore, we can construct the moment-based visual state features, $q$, as follows:
\begin{eqnarray}
{x}_n=a_n{\bar{x}}\\
y_n=a_n{\bar{y}}\\
a_n=Z^*\sqrt{\frac{a^*}{a}}\\
\theta= arctan(1/\rho)
\end{eqnarray}
where   $Z^*$ is the desired normal distance of the camera from the object,  $a^*$ is the desired value of the current area $a$, and $\theta$ is the yaw rate in polar coordinates $(\rho, \theta)$. By setting a as $a=\mu_{20}+\mu_{02}$, the moment-based state equation can be given by
\begin{eqnarray}
\dot{q}&=&L(t)v_c\label{eq:moment_model}\\
L(t)&=&\left[\begin{array}{cccc}
1/Z&0&0&y_n\\
0&1/Z&0&-x_n\\
0&0&1/Z&0\\
0&0&0&-1
\end{array}\right]
\end{eqnarray}
where \( q = [x_n, y_n, a_n, \arctan(1/\rho)]^T \) denotes the selected image feature vector, and the control input 
\( v_c = [v_{xd}, v_{yd}, v_{zd}, \omega_{zd}]^T \) consists of the desired translational velocities 
\( [v_{xd}, v_{yd}, v_{zd}] \) along the three axes and the desired yaw rate \( \omega_{zd} \). The IBVS close-loop control is then given by
\begin{eqnarray}
v_c=-\lambda {L}^{-1}(q-q^*)
\end{eqnarray}
where \( \lambda \) is a user-defined control gain, \( L \) denotes the invertible interaction matrix, and \( q^* \) represents the desired image feature state. However, while computational efficient and simple, this control law does not explicitly account for camera field-of-view (FOV) limitations or control input constraints.

\subsection{Moment-based constrained MPC}
To explicitly incorporate system constraints, the conventional IBVS formulation presented above is reformulated within a Model Predictive Control (MPC) framework. The objective of the proposed moment-based MPC approach is to minimize the visual feature tracking error derived from image moments during the IBVS tracking process. Let \( \mathbf{e}(k) \in \mathbb{R}^{4} \) denote the visual feature error vector at the discrete time step \( k \), constructed from normalized image moment features, including the image centroid and area-related terms. The error vector is defined as
\[
\mathbf{e}(k)=q(k)-q^*(k),
\]
where \( q(k) \) and \( q^*(k) \) represent the current and desired image feature states, respectively. The control input \( \mathbf{u}(k) \in \mathbb{R}^{4} \) corresponds to the camera velocity vector expressed in the camera frame:
\[
\mathbf{u}(k) = [v_x, v_y, v_z, \omega_z]^{\mathsf{T}}.
\]

A discrete-time linear model around the operating point approximates the image system dynamics:
\begin{equation}
\mathbf{e}(k+1) = \mathbf{A}\mathbf{e}(k) + \mathbf{B}\mathbf{u}(k),
\label{eq:linear_model}
\end{equation}
where matrices $\mathbf{A}\in\mathbb{R}^{4\times4}$ and $\mathbf{B}\in\mathbb{R}^{4\times4}$ are obtained from the linearization of the IBVS interaction model around the current operating point.

Specifically, $\mathbf{A}$ is defined as the identity matrix:
\begin{equation}
\mathbf{A} = \mathbf{I}_4
\end{equation}
and $\mathbf{B}$ is obtained from discretizing Eqn \ref{eq:moment_model}:
\begin{equation}
\mathbf{B} = T_s \, L(t) 
\end{equation}

We require the controller to operate at an update frequency of \(f_s = 30\,\mathrm{Hz}\), with the corresponding sampling period \(T_s = 1/f_s\). Under this control assumption, the linearized model provides a sufficiently accurate approximation of the local evolution of the image moment features within the MPC prediction horizon.

For a prediction horizon $N$, the stacked system can be written compactly as:
\begin{equation}
\mathbf{E} = \mathbf{\Phi}\mathbf{e}(k) + \mathbf{\Gamma}\mathbf{U},
\end{equation}
where
\begin{eqnarray}
\mathbf{E} =
\left[\begin{array}{c}
\mathbf{e}(k+1) \\ \mathbf{e}(k+2) \\ \vdots \\ \mathbf{e}(k+N)
\end{array}\right], 
\mathbf{U} =
\left[\begin{array}{c}
\mathbf{u}(k) \\ \mathbf{u}(k+1) \\ \vdots \\ \mathbf{u}(k+N-1)
\end{array}\right].
\end{eqnarray}
The prediction matrices $\mathbf{\Phi}$ and $\mathbf{\Gamma}$ are defined as:
\[
\mathbf{\Phi} =
\begin{bmatrix}
\mathbf{A} \\ \mathbf{A}^2 \\ \vdots \\ \mathbf{A}^N
\end{bmatrix}, \qquad
\mathbf{\Gamma} =
\begin{bmatrix}
\mathbf{B} & 0 & \cdots & 0 \\
\mathbf{A}\mathbf{B} & \mathbf{B} & \cdots & 0 \\
\vdots & \ddots & \ddots & \vdots \\
\mathbf{A}^{N-1}\mathbf{B} & \cdots & \mathbf{A}\mathbf{B} & \mathbf{B}
\end{bmatrix}.
\]

The MPC controller predicts the evolution of the moment-based error $\mathbf{e}(k)$ over the horizon and optimizes the control sequence $\mathbf{U}$ to minimize the deviation from the desired zero-error condition. This formulation enables a direct relationship between the visual error and the optimized camera velocity, providing a principled model-based mechanism for smooth and predictive IBVS control.\\

\noindent\textbf{Objective Function and Online Optimization using Gurobi Solver}\\

The objective of the Moment-based MPC is to minimize the accumulated visual feature error while ensuring smooth control actions and satisfying system constraints. The optimization problem is formulated as a finite-horizon quadratic program (QP) that predicts and optimizes the control sequence $\mathbf{U} = [\mathbf{u}(k), \mathbf{u}(k+1), \ldots, \mathbf{u}(k+N-1)]^{\mathsf{T}}$ over a discrete prediction horizon $N$. The cost function is defined as

\begin{equation}
\begin{aligned}
J = \sum_{i=0}^{N-1} \Big[
&(\mathbf{e}(k+i)-\mathbf{e}_{\mathrm{ref}})^{\mathsf{T}}
\mathbf{Q}
(\mathbf{e}(k+i)-\mathbf{e}_{\mathrm{ref}})
\\
&+
\mathbf{u}(k+i)^{\mathsf{T}}
\mathbf{R}
\mathbf{u}(k+i)
\Big]
\\
&+
(\mathbf{e}(k+N)-\mathbf{e}_{\mathrm{ref}})^{\mathsf{T}}
\mathbf{P}
(\mathbf{e}(k+N)-\mathbf{e}_{\mathrm{ref}})
\label{eq:mpc_cost}
\end{aligned}
\end{equation}

where $\mathbf{Q}, \mathbf{R}$, and $\mathbf{P}$ are the positive semi-definite weighting matrices for state error, control effort, and terminal cost, respectively. In the implemented system, $\mathbf{Q}$ emphasizes the image feature tracking accuracy, while $\mathbf{R}$ penalizes excessive camera velocity changes to ensure smooth flight. The terminal matrix $\mathbf{P}$ ensures convergence near the target. The reference $\mathbf{e}_{\mathrm{ref}}$ corresponds to the desired zero-error moment features $(q^*)$ in the IBVS framework.

The optimization is subject to both state and input constraints that reflect the physical limitations of the UAV:
\begin{align}
\mathbf{e}_{\min} &\leq \mathbf{e}(k+i) \leq \mathbf{e}_{\max}, \quad i = 1, \ldots, N, \\
\mathbf{u}_{\min} &\leq \mathbf{u}(k+i) \leq \mathbf{u}_{\max}, \quad i = 0, \ldots, N-1,
\end{align}
where $\mathbf{u}(k) = [v_x, v_y, v_z, \omega_z]^{\mathsf{T}}$ represents the commanded camera velocity vector. The state constraints correspond to image-space constraints, also referred to as visibility constraints, which ensure that the image features remain within the camera field of view or within a predefined image region. In our implementation, the control bounds are selected according to the maximum translational and rotational velocity limits supported by the PX4 autopilot flight control stack used onboard the UAV. To further ensure closed-loop stability of the MPC controller, an additional terminal constraint is imposed at the final prediction step:
\begin{equation}
|\mathbf{e}(k+N)| \leq \boldsymbol{\varepsilon}_{\mathrm{term}},
\end{equation}
where $\boldsymbol{\varepsilon}_{\mathrm{term}} = [0.5, 0.5, 1.0, 0.5]^{\mathsf{T}}$ represents the allowable terminal error tolerance in each visual dimension. The terminal constraints will help improve the system stability.

At each control iteration, the optimization problem can be written compactly as:
\begin{align}
\min_{\mathbf{U}} \quad & \frac{1}{2} \mathbf{U}^{\mathsf{T}} \mathbf{H} \mathbf{U} + \mathbf{f}^{\mathsf{T}} \mathbf{U}, \\
\text{s.t.} \quad & \mathbf{G}\mathbf{U} \leq \mathbf{w}, \quad \mathbf{E}\mathbf{U} = \mathbf{d},
\label{eq:qp_formulation}
\end{align}
where $\mathbf{H}$ is the Hessian matrix constructed from $\mathbf{Q}$ and $\mathbf{R}$, and $\mathbf{f}$ encodes the linearized prediction model and current error $\mathbf{e}(k)$. The matrices $\mathbf{G}, \mathbf{w}, \mathbf{E}, \mathbf{d}$ describe the inequality and equality constraints derived from system bounds and dynamic equations.

The optimization problem in Eqn. \ref{eq:qp_formulation} is solved online using the Gurobi Optimizer \cite{ref_gurobi}. Gurobi provides efficient sparse quadratic programming solvers that exploit the convexity of Eqn \ref{eq:mpc_cost}. At each sampling instant, the MPC module updates the state vector \(\mathbf{e}(k)\), constructs the optimization matrices, and solves for the optimal control sequence \(\mathbf{U}^\ast\).

The first element of the optimized control sequence,
\begin{equation}
\mathbf{u}(k) = \mathbf{u}_0^\ast,
\end{equation}
is applied to the UAV through the PX4 autopilot for closed-loop velocity control.
\subsection{Moment-based Kalman filter}\label{sect:kalman}
In IBVS, the control law relies on visual features extracted from camera images. However, during quadrotor flight, these visual features may become noisy or temporarily unavailable due to motion blur, illumination variations, occlusions, or rapid target motion. Such degraded measurements can adversely affect the stability and performance of the IBVS controller. Therefore, a Kalman filter is employed to estimate and predict the evolution of the visual feature states from noisy image measurements, thereby improving the robustness and continuity of the visual servoing process.

The objective is to design a filter that tracks the visual feature states and provides reliable estimates even when the corresponding image measurements are temporarily unavailable. Considering the continuous-time model in Eqn. \ref{eq:moment_model}, the discrete-time feature dynamics can be written as

\begin{align}
\mathbf{x}(k) &= \mathbf{A}\mathbf{x}(k-1) + \mathbf{w}(k), \\
\mathbf{z}(k) &= \mathbf{C}\mathbf{x}(k) + \mathbf{v}(k),
\end{align}
where \(\mathbf{x}(k)\) denotes the system state vector, 
\(\mathbf{A} = \mathbf{I}_4\) is the \(4 \times 4\) identity matrix, 
\(\mathbf{z}(k)\) represents the measurement vector, and 
\(\mathbf{C}\) is the observation matrix. The terms 
\(\mathbf{w}(k)\) and \(\mathbf{v}(k)\) denote the process noise and measurement noise, respectively. The noise terms are assumed to follow zero-mean Gaussian distributions:
\[
\mathbf{w}(k) \sim \mathcal{N}(0,\mathbf{Q}), \qquad
\mathbf{v}(k) \sim \mathcal{N}(0,\mathbf{R}),
\]
where \(\mathbf{Q}\) and \(\mathbf{R}\) are the process and measurement noise covariance matrices, respectively.

The Kalman filter consists of a prediction step and a measurement update step. The prediction step is given by
\begin{align}
\hat{x}_{k|k-1} &= \mathbf{A}\hat{x}_{k-1|k-1}, \\
P_{k|k-1} &= \mathbf{A}P_{k-1|k-1}\mathbf{A}^T + Q.
\end{align}

The measurement update step is given by
\begin{align}
\tilde{z}_k &= z_k - \mathbf{C}\hat{x}_{k|k-1}, \\
S_k &= \mathbf{C}P_{k|k-1}\mathbf{C}^T + R, \\
K_k &= P_{k|k-1}\mathbf{C}^TS_k^{-1}, \\
\hat{x}_{k|k} &= \hat{x}_{k|k-1} + K_k\tilde{z}_k, \\
P_{k|k} &= (I-K_k\mathbf{C})P_{k|k-1}.
\end{align}

The estimated feature state \(\hat{x}_{k|k}\) provides a smoothed and continuous estimate of the visual features, improving the robustness of the IBVS controller against noisy or temporarily missing image measurements.

\section{Expermental results}
To validate the effectiveness of the proposed framework, real-world flight experiments were conducted using a quadrotor platform controlled by a Pixhawk flight controller, equipped with an Intel RealSense D435i camera, with the IBVS module executed onboard a Jetson Orin computing platform. The experiments were performed in a motion-capture environment to provide ground-truth pose measurements for evaluation only. 

\begin{figure*}[t]
    \centering
    \begin{subfigure}[b]{0.49\linewidth}
        \centering
        \includegraphics[width=\linewidth,height=4.2cm]{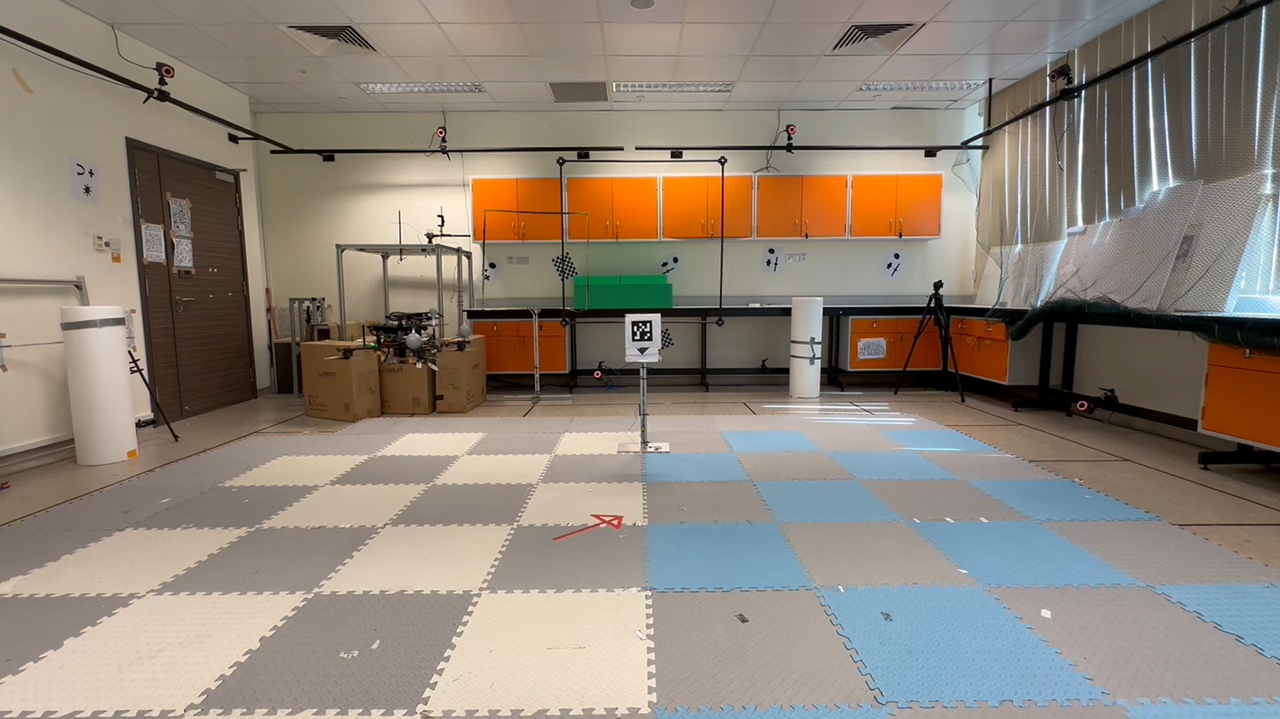}
        \caption{UAV takeoff}
        \label{fig:mpc_takeoff}
    \end{subfigure}
    \begin{subfigure}[b]{0.49\linewidth}
        \centering
        \includegraphics[width=\linewidth,height=4.2cm]{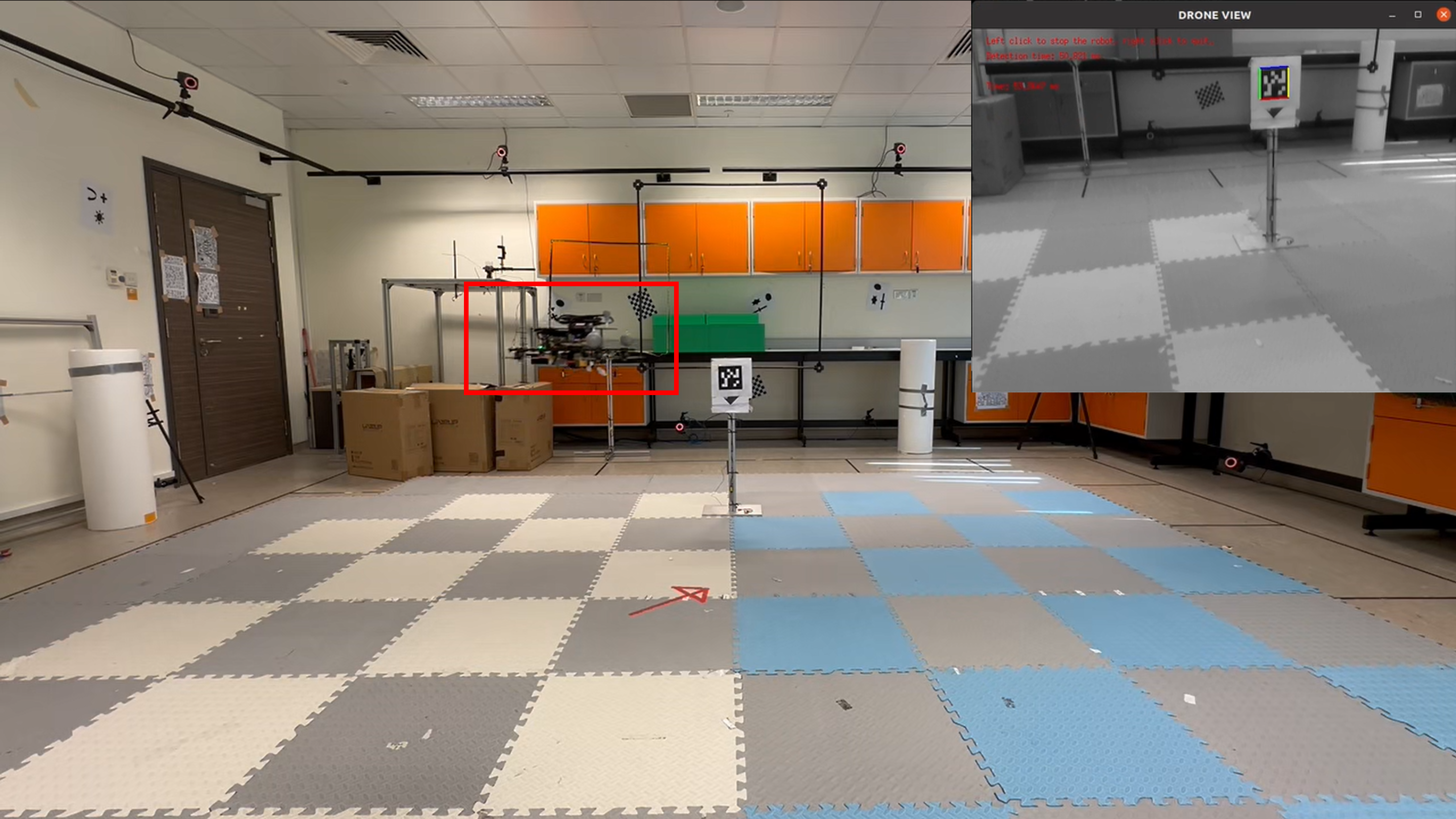}
        \caption{UAV detects the Apriltag}
        \label{fig:mpc_detect}
    \end{subfigure}

    \begin{subfigure}[b]{0.49\linewidth}
        \centering
        \includegraphics[width=\linewidth,height=4.2cm]{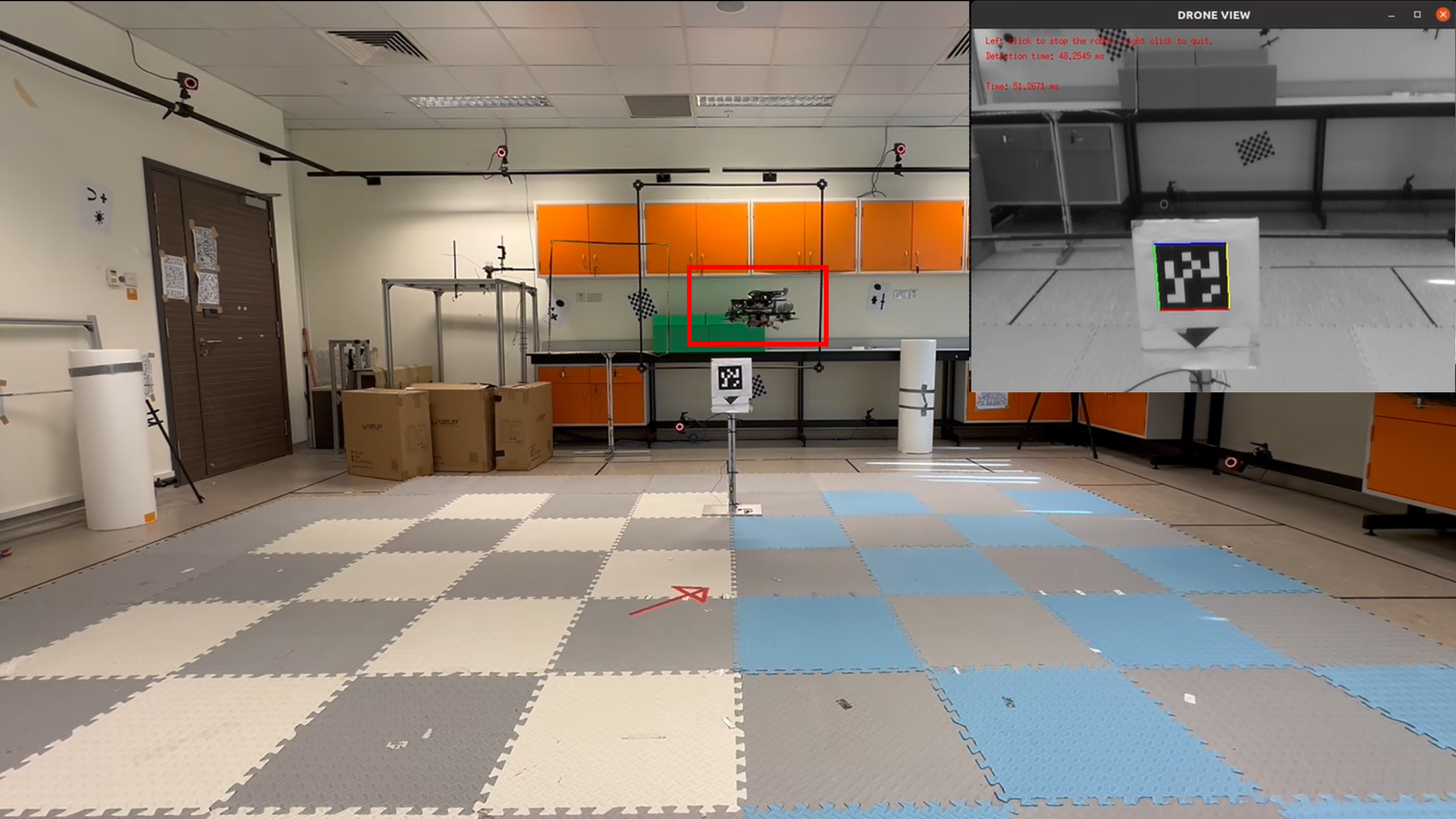}
        \caption{UAV tracks AprilTag}
        \label{fig:mpc_track}
    \end{subfigure}
    \begin{subfigure}[b]{0.49\linewidth}
        \centering
        \includegraphics[width=\linewidth,height=4.2cm]{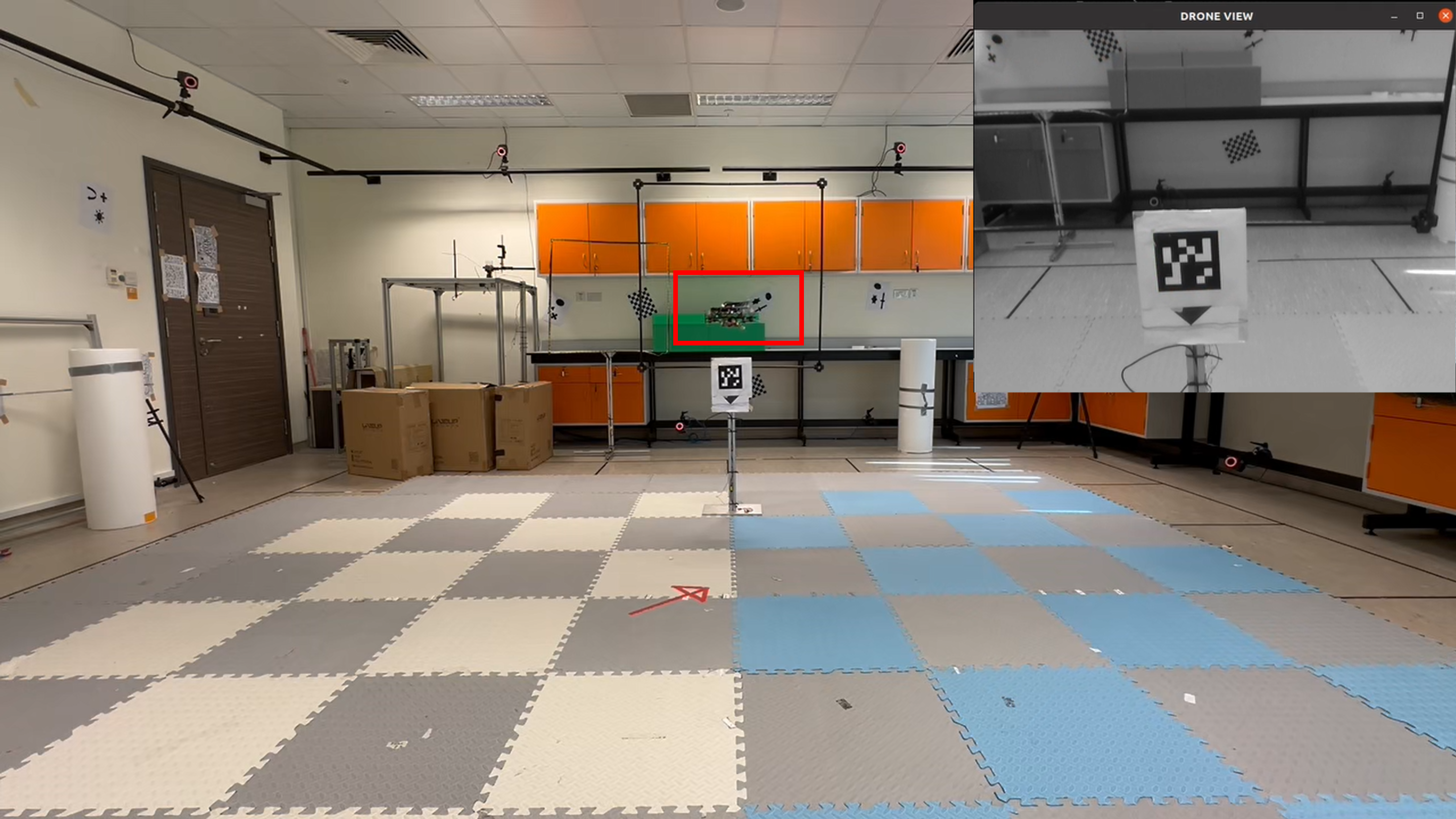}
        \caption{UAV crosses the gate}
        \label{fig:mpc_cross}
    \end{subfigure}

    \caption{A UAV workflow using MPC-based control to track AprilTags and navigate through gates. The drone is highlighted by the red box. The inset images in (b), (c), and (d) show the onboard camera view, where the drone detects and tracks the AprilTag to reach the desired pose.}
    \label{fig:mpc_pipeline}
\end{figure*}

To quantitatively evaluate the proposed method, comparisons were made against conventional IBVS, MPC without constraints (MPC), MPC with input constraints only (MPC1) as proposed in \cite{ref15}, and MPC with both input and state constraints (MPC2). Each method was repeated five times, and the averaged metrics are reported in Table~\ref{tab:rmse_thresholds}. During each trial, the UAV tracked and centered an AprilTag target using onboard visual servoing and MPC executed on the Jetson platform. The overall system workflow is illustrated in Fig.~\ref{fig:mpc_pipeline}.

Three metrics were used for evaluating the tracking accuracy, optimization stability, and convergence performance: the RMSE of the feature tracking error, $e$, the convergence time defined as the time required for the visual error norm \(|\boldsymbol{e}|\) to fall below a threshold of \(0.1\) and a joint RMSE metric was introduced to simultaneously account for tracking accuracy and control smoothness:
\[
\mathrm{RMSE}_{\mathrm{joint}} =
\sqrt{
\mathrm{mean}(e^2) +
\mathrm{mean}(\|\mathbf{v}_e\|^2)
},
\]
where \(e\) denotes the normalized image feature error norm and \(\mathbf{v}_e=[v_x,v_y,v_z]^T\) represents the velocity command generated by the MPC controller. The proposed \(\mathrm{RMSE}_{\mathrm{joint}}\) metric captures the trade-off between visual tracking accuracy and control aggressiveness.

To ensure flight safety, an additional velocity limit of \(1\,\mathrm{m/s}\) was imposed at the IBVS control layer rather than within the MPC optimization problem, preventing the safety constraint from directly affecting the predictive optimization process.

\begin{table}[t]
\centering
\caption{Comparison of tracking accuracy and joint RMSE under different MPC configurations, where MPC denotes MPC without constraints, MPC1 denotes MPC with input constraints, and MPC2 denotes MPC with both state and input constraints.}
\label{tab:rmse_thresholds}
\renewcommand{\arraystretch}{1.1}
\setlength{\tabcolsep}{4pt}
\begin{tabular}{|l|c|c|c|}
\hline
\textbf{Method} & \textbf{Time (s)} & \textbf{RMSE$_\text{error}$} & \textbf{RMSE$_\text{joint}$} \\
\hline
IBVS & 6.588 & 0.7183 & 0.8158 \\
\hline
MPC & 7.633 & 0.7014 & 0.7287 \\
\hline
MPC1~\cite{ref15} & 10.32 & 0.6532 & 0.6785 \\
\hline
MPC2 & 8.176 & 0.6226 & 0.6466 \\
\hline
\end{tabular}
\end{table}

\begin{figure*}[t]
    \centering
    \begin{subfigure}[b]{0.48\linewidth}
        \centering
        \includegraphics[width=\linewidth]{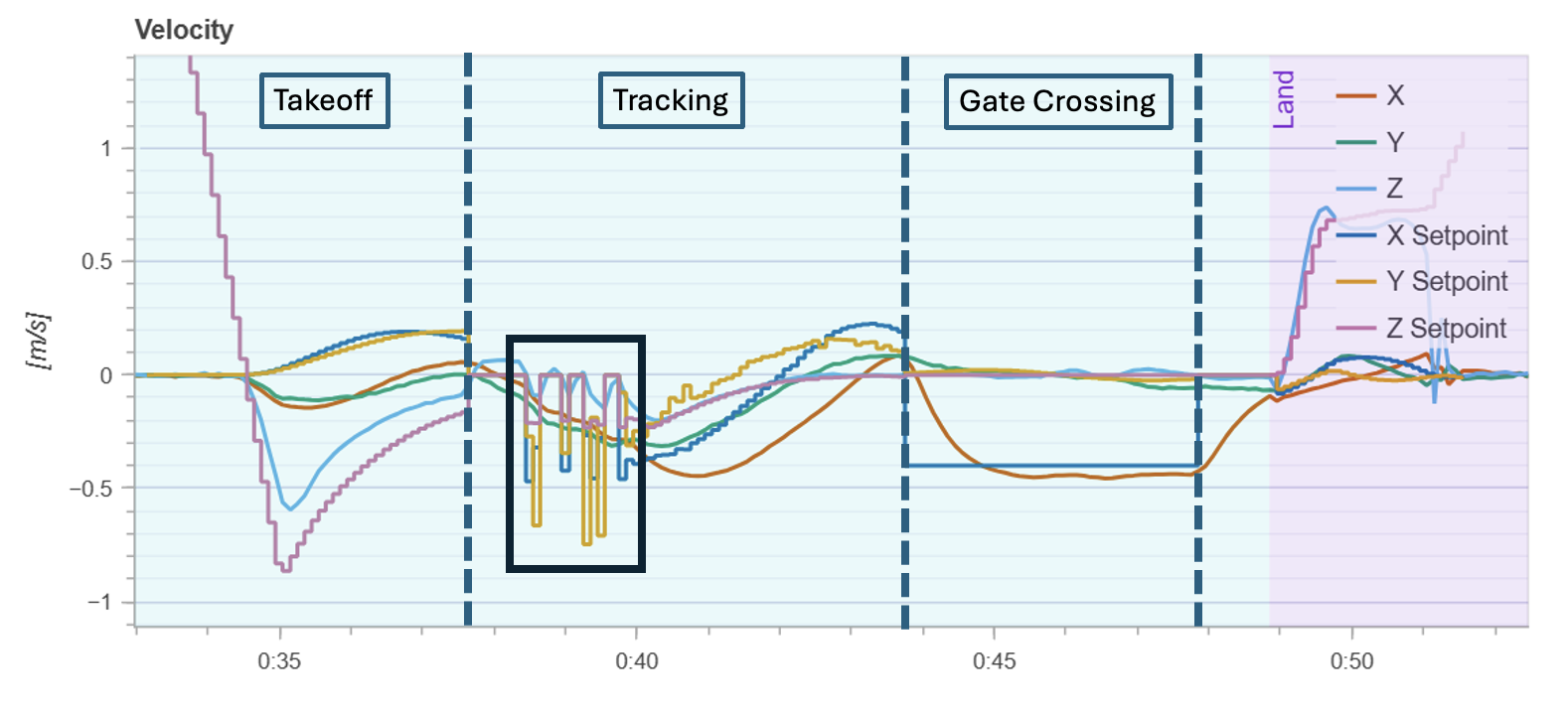}
        \caption{Pure IBVS}
        \label{fig:exp3_no_mpc_pure_ibvs}
    \end{subfigure}
    \hfill
    \begin{subfigure}[b]{0.48\linewidth}
        \centering
        \includegraphics[width=\linewidth]{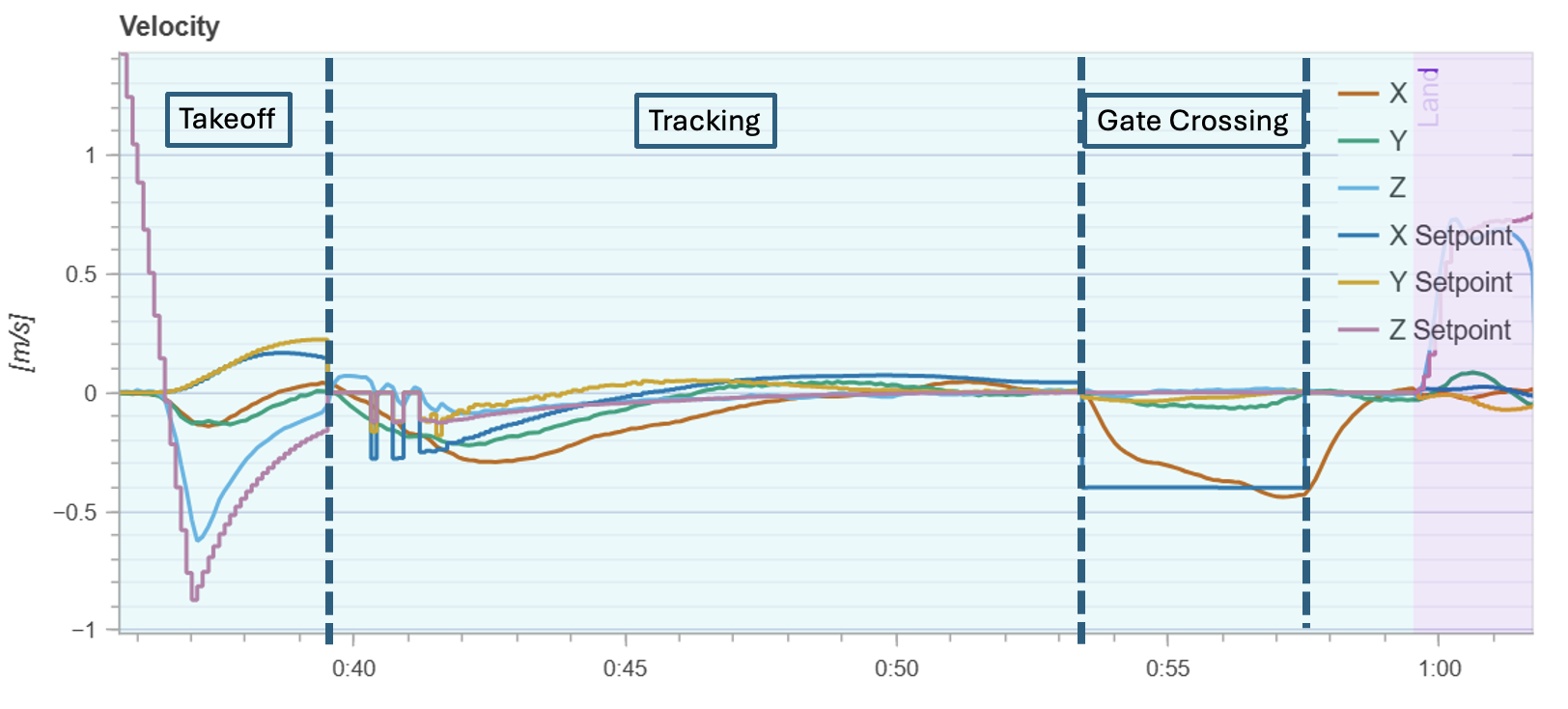}
        \caption{MPC without constraints}
        \label{fig:exp3_mpc_noconstraint}
    \end{subfigure}
    \vspace{0.5em}
    \begin{subfigure}[b]{0.48\linewidth}
        \centering
        \includegraphics[width=\linewidth]{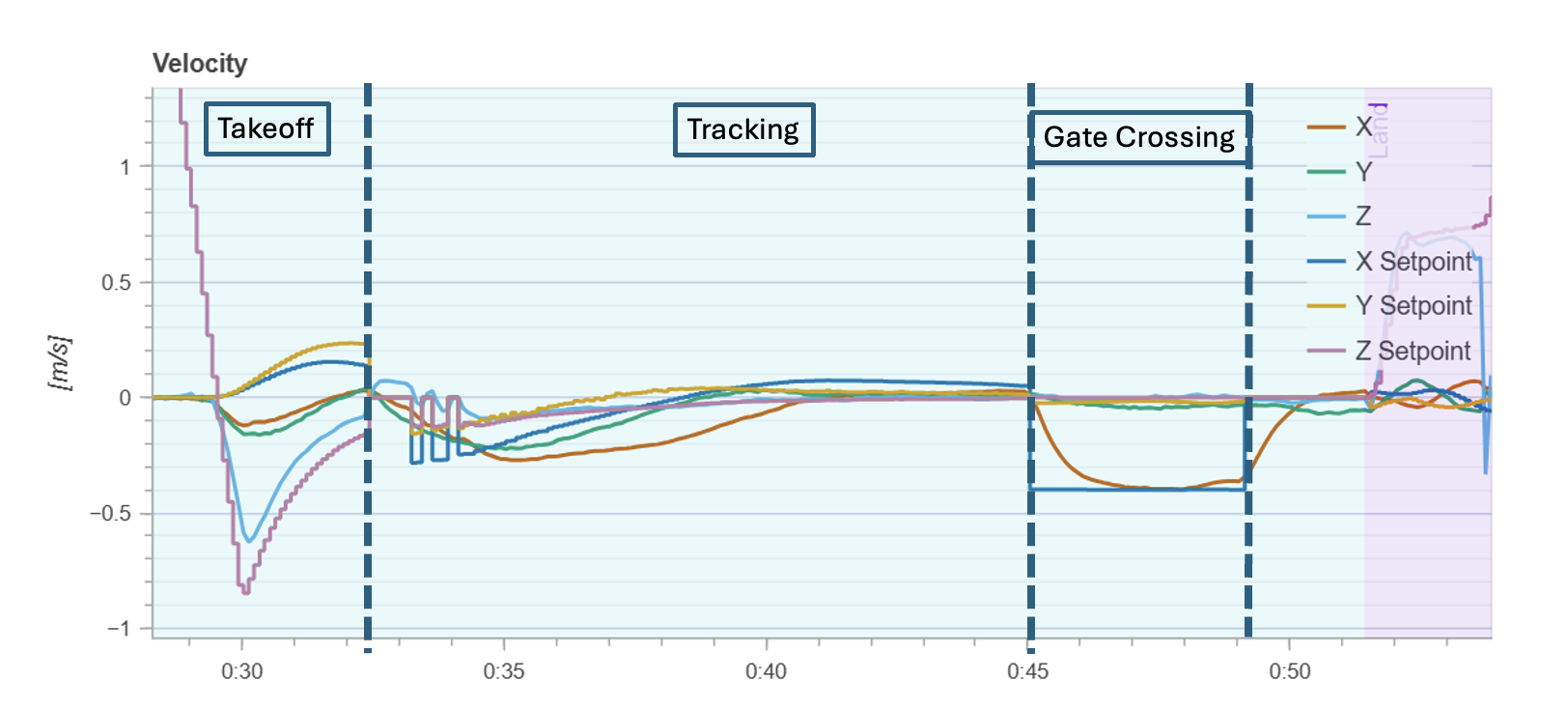}
        \caption{MPC with input constraint}
        \label{fig:exp3_mpc_input_constraint}
    \end{subfigure}
    \hfill
    \begin{subfigure}[b]{0.48\linewidth}
        \centering
        \includegraphics[width=\linewidth]{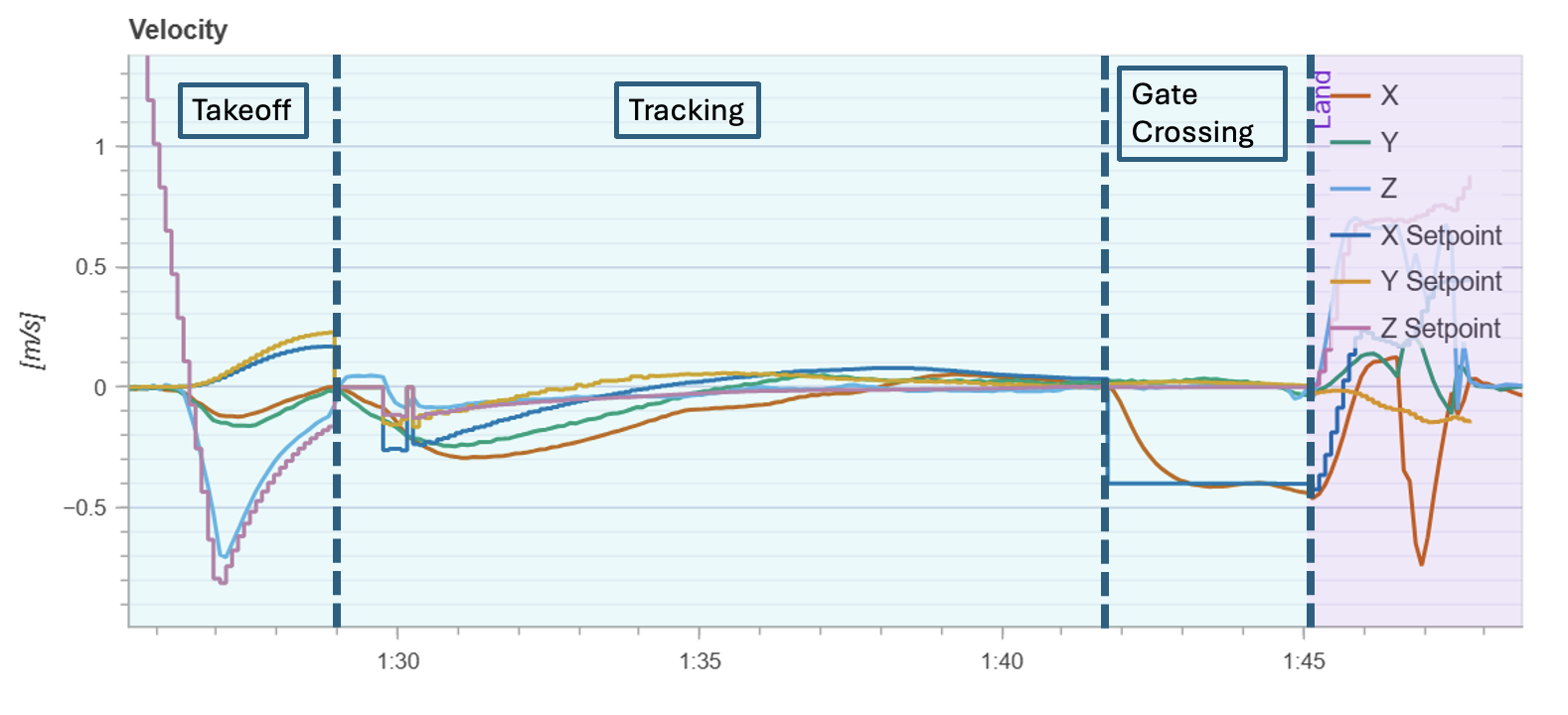}
        \caption{MPC with state and input constraints}
        \label{fig:exp3_mpc_state_input_constraint}
    \end{subfigure}
    \caption{Comparison of UAV velocity tracking performance under different MPC configurations. The figures show current velocities and velocity commands plotted against time. Figure (a) shows an example of oscillatory motion encountered during tracking as marked out by the black box.}
\label{fig:exp3_mpc_velocity_compare}
\end{figure*}

The results show that introducing constraints in the MPC framework improves both stability and accuracy. The fully constrained MPC2
achieved the lowest error among all configurations, with $\mathrm{RMSE}_{error}=0.6226$ and $\mathrm{RMSE}_{joint}=0.6466$, demonstrating smoother motion and better control regularization. Although the unconstrained MPC achieved faster convergence, it occasionally generated aggressive control inputs, reflected by a slightly higher 
joint RMSE. The inclusion of both state and input bounds mitigates this effect, 
producing more stable trajectories.

\begin{figure}[thb]
\centering
\includegraphics[width=1.0\linewidth]{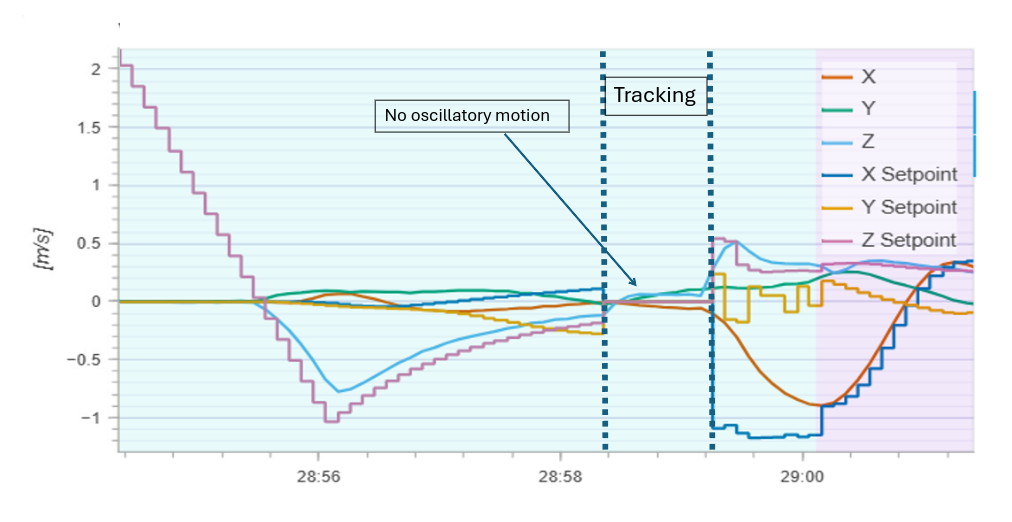}
\caption{IBVS control using MPC with the proposed Kalman filter, demonstrating reduced oscillatory behavior during visual tracking.}
\label{fig:test}
\end{figure}

The experimental results indicate that the generated IBVS commands exhibit noticeable oscillations during the tracking phase, particularly in Figs.~\ref{fig:exp3_no_mpc_pure_ibvs} and \ref{fig:exp3_mpc_noconstraint}. These oscillations are likely caused by measurement noise and intermittent feature loss. Although the inclusion of constrained MPC reduces the oscillation amplitude, residual oscillatory behavior remains. To further improve robustness, the proposed Kalman filter described in Section \ref{sect:kalman} is integrated into the IBVS framework. As shown in Fig.~\ref{fig:test}, the Kalman filter significantly suppresses oscillations during tracking by mitigating the effects of noisy or temporarily missing visual feature measurements.

\section{Conclusion and future works}
In this paper, a terminal constrained MPC-based IBVS framework integrated with a Kalman filter was proposed for quadrotor UAV control. The proposed IBVS formulation is based on image moment features, which alleviates the singularity issues commonly encountered in conventional IBVS methods. To explicitly account for state and control constraints, a terminal MPC framework was developed to ensure stable constrained visual servoing. In addition, a Kalman filter was incorporated to improve robustness against noisy measurements, uncertainties, and temporary feature loss. Real-world experimental results demonstrated that the proposed method enables autonomous gate traversal using only visual features without relying on GPS, while achieving improved performance compared with \cite{ref15}.

Future work will focus on extending the framework to real gate detection without the use of AprilTags, as well as investigating high-speed and agile UAV maneuvers under the proposed visual servoing framework.

%

%

%

\end{document}